\documentclass[conference]{IEEEtran}
\IEEEoverridecommandlockouts
\usepackage{cite}
\usepackage{amsmath,amssymb,amsfonts}
\usepackage{algorithmic}
\usepackage{graphicx}
\usepackage{textcomp}
\usepackage{xcolor}
\def\BibTeX{{\rm B\kern-.05em{\sc i\kern-.025em b}\kern-.08em
    T\kern-.1667em\lower.7ex\hbox{E}\kern-.125emX}}
\begin{document}

\title{Domain Translation of a Soft Robotic Arm using Conditional Cycle Generative Adversarial Network\\
\thanks{We acknowledge the contribution from the Italian National Recovery and Resilience Plan (NRRP), M4C2, funded by the European Union–NextGenerationEU (Project IR0000011, CUP B51E22000150006, "EBRAINS-Italy")}
}

\author{\IEEEauthorblockN{Nilay Kushawaha}
\IEEEauthorblockA{\textit{The BioRobotics Institute} \\
\textit{Scuola Superiore Sant’Anna}\\
Pontedera, Italy \\
nilay.kushawaha@santannapisa.it}
*Corresponding author
~\\
\and
\IEEEauthorblockN{Carlo Alessi}
\IEEEauthorblockA{\textit{Humanoid Sensing and Perception} \\
\textit{Istituto Italiano di Tecnologia}\\
Genoa, Italy \\
carlo.alessi@iit.it}

~\\
\and
\IEEEauthorblockN{Lorenzo Fruzzetti}
\IEEEauthorblockA{\textit{Neuroscience Institute} \\
\textit{CNR - Consiglio Nazionale Ricerche}\\
Pisa, Italy \\
lorenzofruzzetti1@gmail.com}
~\\
\and

\hspace{7cm}\IEEEauthorblockN{Egidio Falotico}
\IEEEauthorblockA{\hspace{7cm}\textit{The BioRobotics Institute} \\
\hspace{7cm}\textit{Scuola Superiore Sant’Anna}\\
\hspace{7cm}Pontedera, Italy \\
\hspace{7cm}egidio.falotico@santannapisa.it}

}

\maketitle

\begin{abstract}
Deep learning provides a powerful method for modeling the dynamics of soft robots, offering advantages over traditional analytical approaches that require precise knowledge of the robot’s structure, material properties, and other physical characteristics. Given the inherent complexity and non-linearity of these systems, extracting such details can be challenging. The mappings learned in one domain cannot be directly transferred to another domain with different physical properties. This challenge is particularly relevant for soft robots, as their materials gradually degrade over time. In this paper, we introduce a domain translation framework based on a conditional cycle generative adversarial network (CCGAN) to enable knowledge transfer from a source domain to a target domain. Specifically, we employ a dynamic learning approach to adapt a pose controller trained in a standard simulation environment to a domain with tenfold increased viscosity. Our model learns from input pressure signals conditioned on corresponding end-effector positions and orientations in both domains. We evaluate our approach through trajectory-tracking experiments across five distinct shapes and further assess its robustness under noise perturbations and periodicity tests. The results demonstrate that CCGAN-GP effectively facilitates cross-domain skill transfer, paving the way for more adaptable and generalizable soft robotic controllers.
\end{abstract}


\section{Introduction}
The modeling and control of continuum and soft robotic arms remain challenging due to their hyper-redundancy, complex dynamics, and the inherent non-linear properties of soft materials \cite{rus2015design}, \cite{kushawaha2025adaptive}. These manipulators typically feature a single, flexible backbone that bends and twists in response to actuation \cite{falotico2024learning}. Actuation can be distributed, such as in pneumatic systems, or localized, as seen in cable-driven mechanisms. Various modeling approaches \cite{armanini2023soft} have been explored, leading to the development of both model-based and model-free control strategies. However, the reliability of model-based methods decreases when the robot interacts with unpredictable environments or encounters substantial external forces.

Learning-based techniques provide an alternative for approximating the kinematics \cite{melingui2015adaptive}, \cite{kalidindi2019cerebellum} and dynamics \cite{thuruthel2017learning} of soft robots, eliminating the need for explicit analytical models. These methods are particularly beneficial for tasks requiring adaptation to unknown forces and torques or maneuvering through constrained spaces, challenges that are more straightforward in rigid-link robots but add complexity in soft robotics \cite{george2017learning}, \cite{alessi2024pushing}, \cite{bianchi2024softsling}. However, transferring learned dynamic behaviors from one domain to another can be difficult due to variations in physical properties \cite{alessi2023learning}. For instance, soft robots may experience morphological degradation due to environmental conditions or material aging \cite{hutchinson1995physical}. Training a controller solely on babbling data from a target domain often embeds domain-specific constraints, leading to performance degradation when applied to trajectories from a slightly different domain. To address this, it is crucial to establish an effective coupling between the domains to enable skill transfer while minimizing performance loss.

In this work, we address this challenge by employing a cycle-consistent generative adversarial network (Cycle-GAN) \cite{zhu2017unpaired} to perform domain translation, facilitating the transfer of skills learned in a standard simulation environment to a constrained domain characterized by a 10x increase in viscosity of the environment. Specifically, a Conditional Cycle GAN with Gradient Penalty (CCGAN-GP) \cite{gulrajani2017improved} is introduced to learn a joint actuation mapping between the two simulation domains, enabling efficient and reliable skill transfer. To the best of our knowledge, we are the first to address the issue of domain transfer for a soft robot by applying the concept of cycle GANs. The main contributions of this work are summarized as follows: 

\textbf{1.} \textbf{Domain translation in soft robotics.} We propose a novel domain translation approach for soft robots using a CCGAN-GP-based inverse dynamics algorithm, enabling the transfer of a pose (position and orientation) controller between two unpaired simulation domains. This contribution is the first demonstration of domain transfer for controlling soft robots.

\textbf{2.} \textbf{Validation and benchmark.} We validate the proposed framework on trajectory-tracking tasks across five distinct shapes. Additionally, we perform a noise robustness test by introducing perturbations to the input signals before feeding them into the generator model and conduct a periodicity test to evaluate the stochasticity of the generator’s predictions.
\section{Related Work}
\begin{figure}[t]
	\centering
	\includegraphics[]{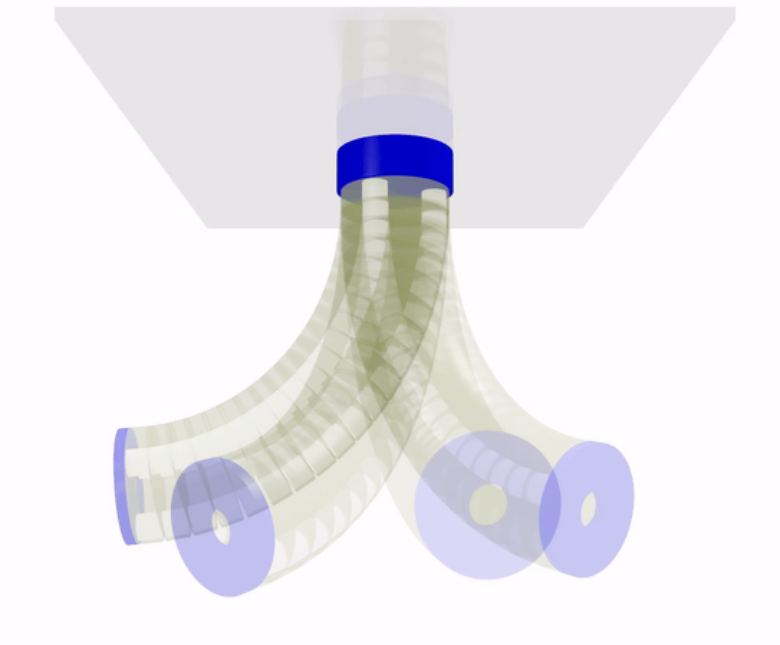}
	\caption{Single module soft robotic arm affixed on a rigid surface, allowing its free end to move and actuate around the complete workspace.}
	\label{fig:robot_img}
\end{figure}

\subsection{Domain transfer with Cycle-GAN}
Cycle-GAN has been widely applied to domain transfer tasks involving image datasets \cite{liu2023digital}, \cite{shi2023sim}, \cite{chen2022bidirectional}, \cite{yang2020regularized}, whereas its use for time series signal transformation remains relatively underexplored. Notably, \cite{luleci2023cyclegan} leveraged Cycle-GAN for domain translation between undamaged and damaged acceleration signals in structural health monitoring. Similarly, \cite{liao2024novel}, \cite{yu2025novel}, and \cite{chen2024cyclegan} employed Cycle-GAN to synthesize faulty rolling bearing signals, mitigating class imbalance in classifier training. In a different context, \cite{jia2020gan} employed Cycle-GAN to perform language-independent emotion transfer, enabling the translation of emotions learned in one language to another without the need for paired training samples. Furthermore, \cite{basak2024novel} applied Cycle-GAN to reconstruct fetal electrocardiogram (fECG) signals from maternal electrocardiogram (mECG) data while preserving their morphological features. Additionally, Xu et al. \cite{xu2025enhancing} utilized Cycle-GAN for denoising laser sensor signals, improving measurement accuracy. Despite these advancements, no prior research has investigated the application of Cycle-GAN for domain transfer in soft robot control.

\subsection{Learning-based Soft Robot Control}
Learning-based methods have been extensively utilized in the literature to control soft continuum robots, primarily through supervised training or reinforcement learning (RL) algorithms \cite{falotico2024learning}. These approaches are commonly employed to model the robot's dynamics based on actuation inputs and the resulting position and orientation \cite{pagliarani2025softtex}. For instance, \cite{george2017learning} adopted a model-free strategy to learn the inverse kinematics of a soft continuum robot while incorporating end-effector feedback for reaching and tracking tasks. Similarly, \cite{chen2022bidirectional} leveraged a bi-directional long short-term memory (LSTM) network to develop a controller for a modular soft robot. In another study, \cite{alessi2024pushing} employed a proximal policy optimization (PPO)-based algorithm to learn a closed-loop controller for pushing tasks. \cite{bianchi2023softoss} utilized an open-loop RL approach to generate an actuation pattern for a tossing task based on the target position. Additionally, \cite{nazeer2024rl} integrated RL with behavior cloning to achieve high-precision control of soft robots in the presence of external disturbances for reaching tasks. However, these methods have been applied only within a single domain and do not consider domain transferability.

\begin{figure}[t]
	\centering
	\includegraphics[width=\columnwidth]{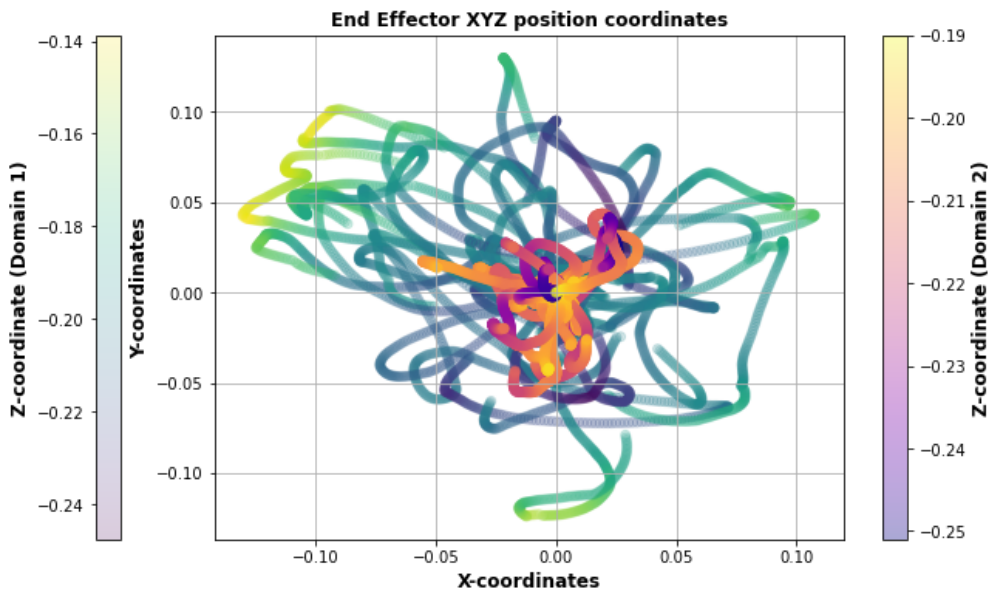}
	\caption{The plot shows the babbling data for the two domains $D_s$ (shown in green) and $D_c$ (shown in violet) with the end effector position coordinates (all axes are measured in metres). Note: we only display half of the trajectories here for simplicity.}
	\label{fig:domain_data}
\end{figure}
\section{Experimental Setup}

\subsection{Simulation Environment}
The simulation environment for this study is based on the soft robotic arm model described in \cite{alessi2023ablation}. This model represents the soft robotic arm as a Cosserat rod with a constant cross-section and homogeneous material properties, extending the Cosserat rod theory from \cite{gazzola2018forward} to incorporate pneumatic actuation. The model includes a simplified representation of the pneumatic actuation system, capturing the time-dependent response of proportional pressure-controlled electronic valves, enabling the soft arm to stretch and bend its backbone.  

The soft robotic arm, as shown in Figure \ref{fig:robot_img} has a cross-section radius of approximately 30 mm, a total length of around 202 mm, and a weight of about 183 g. The pneumatic chambers are roughly 180 mm long, with actuators distributed axially at a radial distance of 20 mm from the cross-section centroid and spaced evenly at 120$^{\circ}$ intervals around the center. The arm is influenced by gravity and can experience viscous forces applied to its free end, which can be integrated into its body dynamics.  

In this study, we use a single module of the arm operating at 10 Hz across both domains. The simulation and environmental parameters remain consistent between the two domains, except for the viscous force, where the target domain \(D_c\) exhibits higher viscosity than the source domain \(D_s\).

\subsection{Data Collection \& Pre-processing}

The soft robotic arm operates in two distinct simulated environments: Domain 1 (\(D_s\)) and Domain 2 (\(D_c\)). Domain 1 represents a \textit{standard} simulation environment, while Domain 2 introduces a \textit{constrained} setting by increasing the medium’s viscosity to 10 times that of \(D_s\), effectively simulating higher friction. To collect data in these environments, we use exploratory "babbling" movements, where pseudo-random pressure values are applied to all three pneumatic chambers, ensuring comprehensive coverage of the workspace. The actuation and sampling frequencies are maintained at 10 Hz in both the simulation domains.  

As depicted in Figure \ref{fig:domain_data}, we generate 50 random trajectories per domain, each lasting 60 seconds. This dataset is used to train both the forward and inverse models of the soft robot. The plot illustrates the XYZ coordinates of the end-effector positions, while the dataset also includes the \(d_3\) component \cite{alessi2023ablation}, which encodes orientation information in all three directions. After each trajectory, the simulation resets, allowing the arm to start the next movement from an initial resting position. To assess model performance during training, we generate a separate validation set consisting of three additional random trajectories, evaluated every \(N\) steps.  

In total, the training dataset comprises 15,000 data points, with three feature columns representing actuation values and six target columns corresponding to the end effector's position and orientation. Orientations are computed using the cross-product of two vectors constructed at the base of the end effector. All actuation and task-space values are normalized to a range of 0 to 1. Actuation values are scaled by dividing each pressure reading by the maximum achievable pressure of 3.5 bars, while task-space values are normalized using synthetic data spanning the complete workspace in both domains. Additional custom trajectories are generated similarly but use precise actuation signals to trace specific shapes.

\section{Methodology}
In this section, we introduce the proposed CCGAN algorithm and demonstrate its capability to model the dynamics of a soft robotic arm across two distinct, unpaired domains. We design an end-to-end training pipeline for CCGAN to learn the inverse dynamics of the soft arm by integrating it with the robot’s forward model. During the training phase, instead of relying on the simulator directly, we rather train a forward model on the babbling data to approximate the simulated arm. This is due to the simulator’s limited speed during batch-wise training and lack of parallelization which significantly effects training time in a closed-loop setup. The implementation details of each component are outlined below:

\begin{figure}[t]
	\centering
	\includegraphics[]{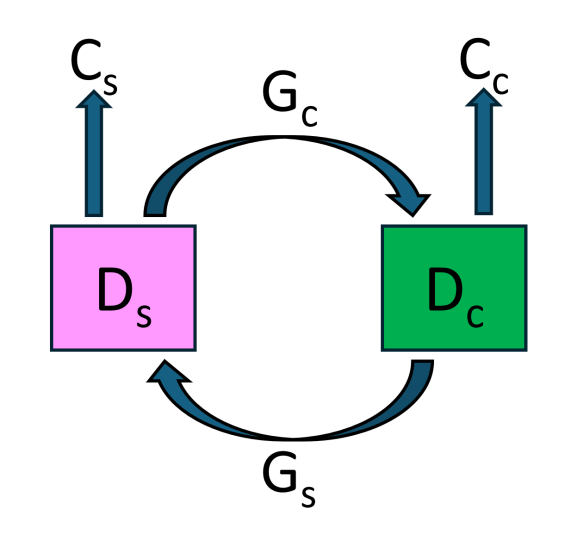}
	\caption{Our framework contains two mapping functions that translates from $D_s$ to $D_c$ and vice-versa. The discriminator $C_c$ encourages $G_c$ to translate $D_s$ into outputs indistinguishable from domain $D_c$, and vice-versa for $C_s$ and $G_s$.  }
	\label{fig:mapping_figure}
\end{figure}

\subsection{Problem Statement}
The goal of our framework is to learn a mapping function between the two domains (as shown in Figure \ref{fig:mapping_figure}): domain 1 (source domain, $D_s$) and domain 2 (constrained domain, $D_c$) given training samples $\{d^i_s\}^N_{i=1}$ where $d^i_s \in D_s \forall D_s = \{\tau^i_s, X^i_s, \phi^i_s\}$ and $\{d^i_c\}^N_{i=1}$ where $d^i_c \in D_c \forall D_c = \{\tau^i_c, X^i_c, \phi^i_c\}$. The proposed framework contains two generator functions $G_s : D_c \longrightarrow D_s$ and $G_c : D_s \longrightarrow D_c$ and the associated critics $C_s, C_c$. These critics guide the generators to produce realistic actuation values that align with the characteristics of the original babbling dataset.

Critic \(C_c\) ensures that \(G_c\) accurately translates actuation values from \(D_s\) to \(D_c\) such that, when processed through the forward model of \(D_c\), the resulting trajectory closely follows the one observed in \(D_s\). During testing, given actuation values for an unknown shape in \(D_s\), we can use \(G_c\) to generate the corresponding actuation values required to produce an equivalent trajectory in \(D_c\). This approach is particularly beneficial when \(D_c\) represents a constrained environment with friction or other dissipative forces, where directly transferring a trajectory from \(D_s\) to \(D_c\) is impractical without jointly training both domains. In such cases, training a standard inverse model solely on the babbling data from \(D_c\) is insufficient for trajectory transfer from \(D_s\) due to differences in the physical properties of the target domain.

\subsection{Working \& Model Architecture}
The CCGAN is a specialized form of cycle generative adversarial network (Cycle-GAN) where additional conditioning terms are introduced to integrate a mapping between two distinct input distributions. One of the main advantages of Cycle-GAN, as previously mentioned, is its capacity to train on unpaired data while learning both the forward mapping from the source domain $(D_s)$ to the target domain $D_c$ and the inverse mapping back to the source domain.

To stabilize the training of the Cycle-GAN, we follow the techniques outlined in \cite{gulrajani2017improved}, particularly by applying a gradient penalty to enforce the Lipschitz constraint on the critic rather than using weight clipping, as was proposed in the original Cycle-GAN paper \cite{zhu2017unpaired}. Additionally, we use the wasserstein distance for adversarial training, which mitigates the vanishing gradient problem that can occur with Jensen-Shannon (JS) divergence as the critic's output saturates during training.

The complete architecture comprises two generators, two critics, and two distinct forward dynamics models for each domain. The architecture is trained in a hybrid approach that combines adversarial training with the error obtained from the forward model as a feedback. Further details on each model are provided below.

\subsubsection{Forward Dynamics Model}
The purpose of the forward model is to serve as a decent approximation of the robot's forward dynamics mapping for the training phase of the GAN-based controller. It learns a mapping from the actuation space ($\tau$) to the task space ($X$). Specifically, the forward model takes as input the end effector's previous position $X_{i-1}$, orientation $\phi_{i-1}$, as well as the current actuation \textit{$\tau_{i}$} and outputs the end effector's position $X_{i+1}$ and orientation $\phi_{i+1}$ for the next time step, where $X_{i-1/i+1}, \phi_{i-1/i+1} \in \textbf{R}^{6}$ and $\tau_{i} \in \textbf{R}^{3}$. This approach allows the model to capture the system’s dynamics by incorporating both pressure and position information over time.

Recurrent neural networks (RNNs), particularly LSTM architectures, are well-suited for learning temporal dependencies, as demonstarted in \cite{pique2022controlling}. Therefore, we use a single-layer LSTM with 64 neurons to model the robot’s dynamics, followed by a fully connected layer with 50 neurons, a tanh activation function, and a final linear layer with 6 neurons to predict the end-effector’s position and orientation.

A separate forward model is trained for each of the two domains, using the same architecture across both. The models are optimized using the adam optimizer with a weight decay of $10^{-4}$ to prevent overfitting, and mean squared error (MSE) as the loss function to minimize the difference between predicted and actual end-effector poses. Training is conducted using the same babbling dataset as the inverse model, consisting of 50 different trajectories, as described in the previous section.

\subsubsection{Generator Model}
The proposed generator network is designed to work as the inverse dynamics model of the simulated soft robotic arm to capture its physical properties. The generator model used in this study is a variant of traditional GAN but with domain interchangeability capabilities to learn the mapping from the task space to the respective actuation space. The generator model takes as input the current actuation ($\tau^{D_s}_i$) of domain 1 followed by the current end effector position, orientation ($X^{D_s}_i, \phi^{D_s}_i$) as a condition and predicts the corresponding actuation in domain 2 ($\tau^{D_c}_i$) followed by the conditional part ($X^{D_c}_i, \phi^{D_c}_i$). The condition part helps the generator to have a virtual conditional mapping between the actuation space and the task space of the dataset. To provide this conditional information, we simply concatenate the actuation values with the respective end-effector positions and orientations for each domain.

The generator model uses a similar LSTM-based architecture as described above, specifically, it consists of two LSTM layers with 32 neurons each, followed by a fully connected layer with 128 neurons and a ReLU activation function, and a final output layer with 9 neurons and a linear activation function. While training the CCGAN-GP architecture we use two identical generators for the two domains as discussed in more detail in the next section. The models are optimized using a combination of different loss functions (discussed in Section \ref{losses_sec}) and trained with the adamW optimizer \cite{loshchilov2017decoupled} (\(\beta_1 = 0.5\), \(\beta_2 = 0.999\)). AdamW is chosen for its ability to decouple weight decay from gradient updates, preventing interference with the effective learning rate and enhancing stability. It also includes an L2 regularization term to penalize large weights.  

Additionally, the model incorporates an adaptive time-stretching phase, where the generator’s output for the constrained domain \(D_c\) is interpolated by a variable factor to align with the source domain \(D_s\). This phase (elaborated in Section \ref{train_gan}) ensures that the generator for the constrained domain (\(G_c\)) covers the same workspace as the source domain.

\begin{figure}[t]
	\centering
	\includegraphics[width=\columnwidth]{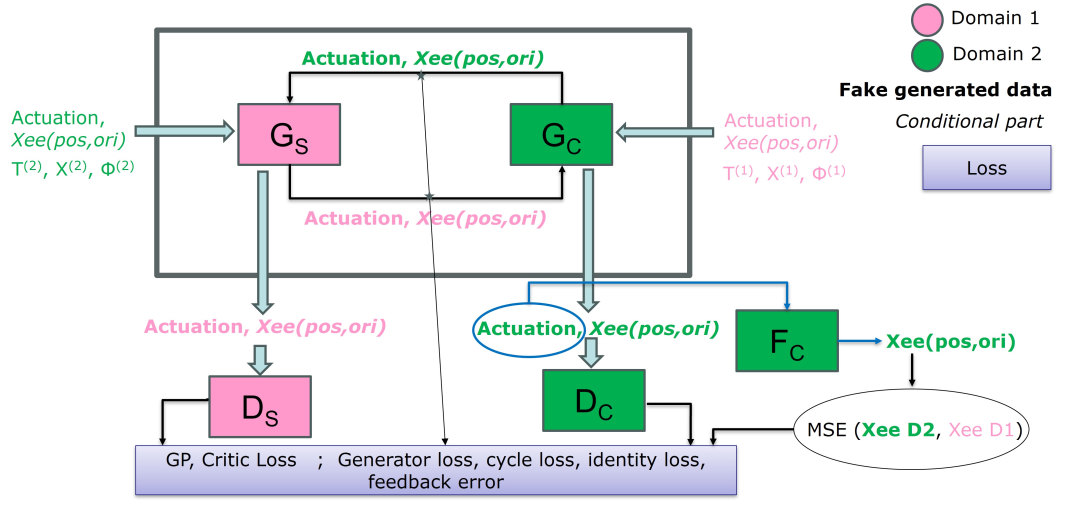}
	\caption{The CCGAN-GP architecture consists of two generator and two discriminator networks, and one forward dynamics model for the constrained domain ($D_c$). }
	\label{fig:architecture}
\end{figure}

\subsubsection{Critic Model}
The output from the generator is evaluated by the critic to assess the quality of the synthetic data. Specifically, the critic attempts to distinguish whether the input data is real (from the original domain) or fake (synthetic data generated by the opposite domain’s generator). For each generator, there is a corresponding critic model to evaluate the quality of the generated data. The critic architecture consists of five 1D-convolutional layers followed by an instance normalizer and leaky relu activation function. It takes as input the actuation values $\in \textbf{R}^{3}$, conditioned by the end effector pose $\in \textbf{R}^{6}$ for both real and fake data. The output is a real-valued scalar, where higher values imply "more real" and lower values imply "more fake." The critic's goal is to assign higher scores to real data samples and lower scores to generated (fake) samples. The difference in these scores is used to approximate the wasserstein-1 distance between the real and generated data distributions. A gradient penalty term is added to the critic's output to enforce a Lipschitz constraint \cite{gulrajani2017improved}, penalizing gradients that deviate from unit value. This penalty helps stabilize the training process by preventing excessive optimization of the critic, ensuring the wasserstein distance approximation remains valid.

\begin{figure}[t]
	\centering
	\includegraphics[width=\columnwidth]{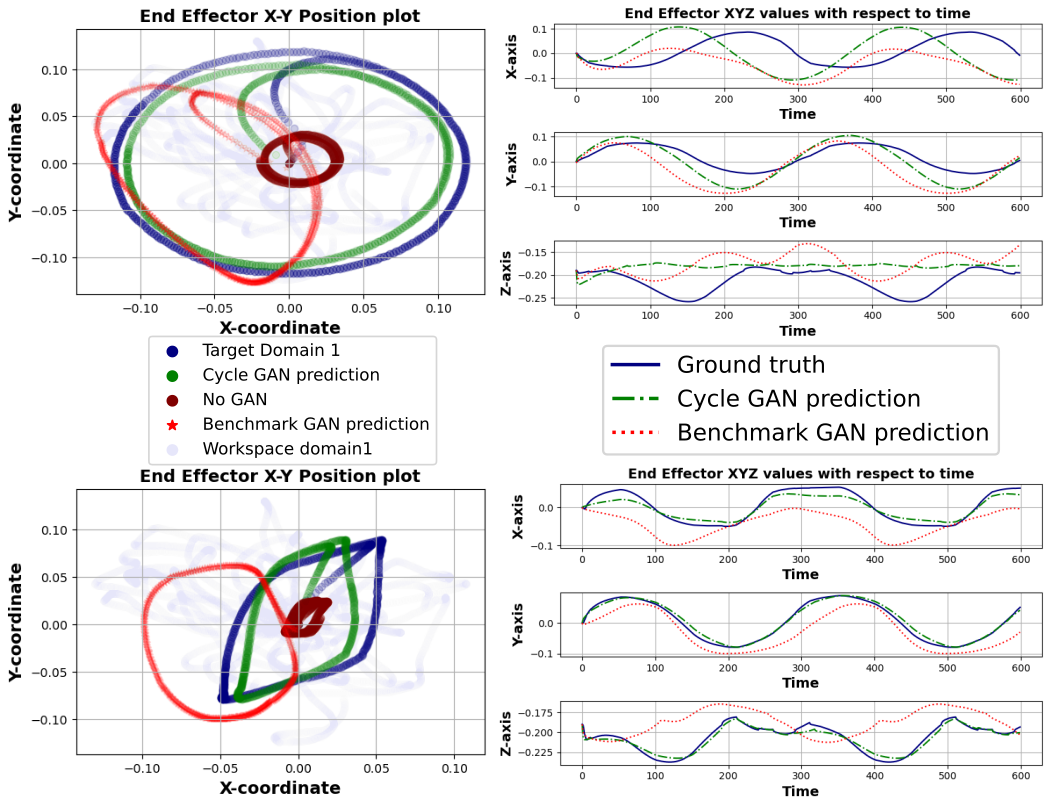}
	\caption{End effector position and orientation error for the circular and rectangular shape, left plot shows the planar X-Y trajectory whereas the right plot highlights the XYZ trajectory with respect to time.}
	\label{fig:circle_rect}
\end{figure}

\subsection{Loss Function Formulation}\label{losses_sec}
The complete loss function for training the CCGAN-GP consists of two different parts total generator loss and total discriminator loss. As discussed in the previous section the total discriminator loss is a combination of wasserstein distance loss and gradient penalty loss. On the other hand, the generator loss is a combination of four distinct terms: adversarial loss, cycle loss, identity loss, and feedback loss. Each of these loss terms is explained in more detail below for better clarity:

\begin{equation}
    \begin{aligned}
    \mathcal{L}^{Adversarial}_{(G_s, G_c)} = -\big( \mathbb{E}_{d^{i}_{c} \forall d^{i}_{c} \in D_c} [C_s (G_s (d^i_c))]  \\+ \mathbb{E}_{d^{i}_{s} \forall d^{i}_{s} \in D_s} [C_c (G_c (d^i_s))] \big)          
    \end{aligned}
\end{equation}

\begin{equation}
    \begin{aligned}
    \mathcal{L}^{Cycle}_{(G_s, G_c)} = \mathbb{E}_{d^{i}_{s} \forall d^{i}_{s} \in D_s} \left\| G_s (G_c (d^i_s)) - d^{i}_s \right\|_s  \\+ \mathbb{E}_{d^{i}_{c} \forall d^{i}_{c} \in D_c} \left\| G_c (G_s (d^i_c)) - d^{i}_c \right\|_s          
    \end{aligned}
\end{equation}

\begin{equation}
    \begin{aligned}
    \mathcal{L}^{Identity}_{(G_s, G_c)} = \mathbb{E}_{d^{i}_{s} \forall d^{i}_{s} \in D_s} \left\| G_s (d^i_s) - d^{i}_s \right\|_s  \\+ \mathbb{E}_{d^{i}_{c} \forall d^{i}_{c} \in D_c} \left\| G_c (d^i_c)) - d^{i}_c \right\|_s          
    \end{aligned}
\end{equation}

\begin{equation}
    \begin{aligned}
    \mathcal{L}^{Feedback}_{G_c} = \mathbb{E}_{d^{i}_{s} \forall d^{i}_{s} \in D_s} \big( F_c (\tau^i_{G_c}) - [X^{d_c}, \phi^{d_c}])\big)^2          
    \end{aligned}
\end{equation}

Finally, the total loss function becomes:

\begin{equation}
    \mathcal{L}^{Total}_{G_s, G_c} = \mathcal{L}^{Adversarial}_{(G_s, G_c)} + \lambda_c\mathcal{L}^{Cycle}_{(G_s, G_c)} + \mathcal{L}^{Identity}_{(G_s, G_c)} +  \lambda_f\mathcal{L}^{Feedback}_{G_c}         
\end{equation}

where $\tau^i_{G_c}$ denotes the actuation predictions made by the $G_c$ model, $\lambda_c$ and $\lambda_f$ denotes hyperparameter that controls the importance of the respective terms in the overall objective function. The adversarial loss establishes a min-max game between the generator and discriminator networks, encouraging the generator to produce realistic outputs. The cycle loss enables the generator to learn bidirectional domain translation between the source and target domains. The identity loss ensures that the generator can maintain consistency by correctly mapping inputs when $G_s$ is provided with data from the same domain, $D_s$. Lastly, the feedback loss enhances the generator's ability to learn the robot's dynamics by utilizing the task-space error

\begin{figure}[t]
	\centering
	\includegraphics[width=\columnwidth]{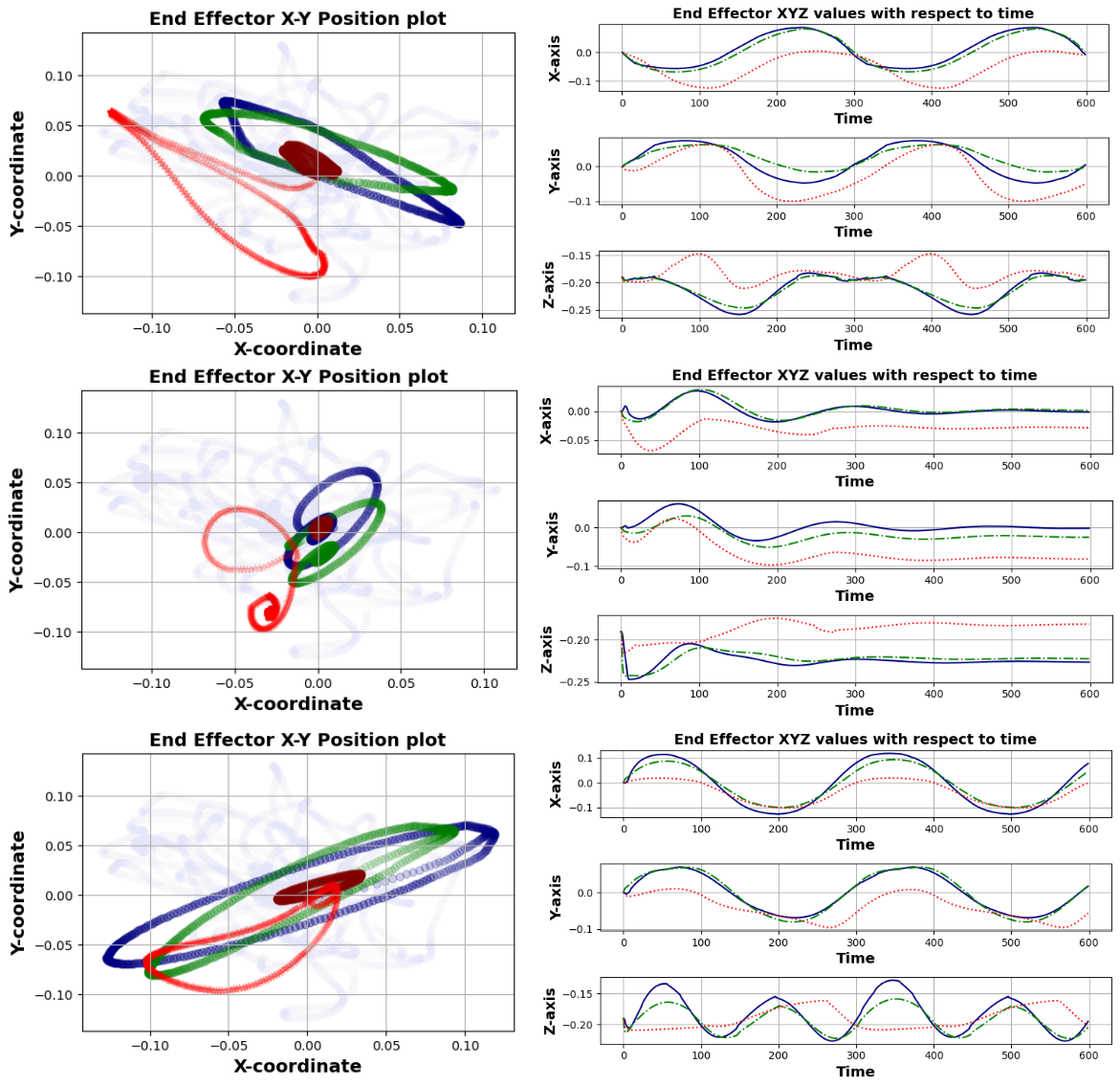}
	\caption{End effector position and orientation error for the kite, damped circle and an elliptical shape. All these shapes are generated by actuating the robot for 60 seconds at 10 Hz.}
	\label{fig:circle_rect2}
\end{figure}

\begin{figure*}
	\centering
	\includegraphics[]{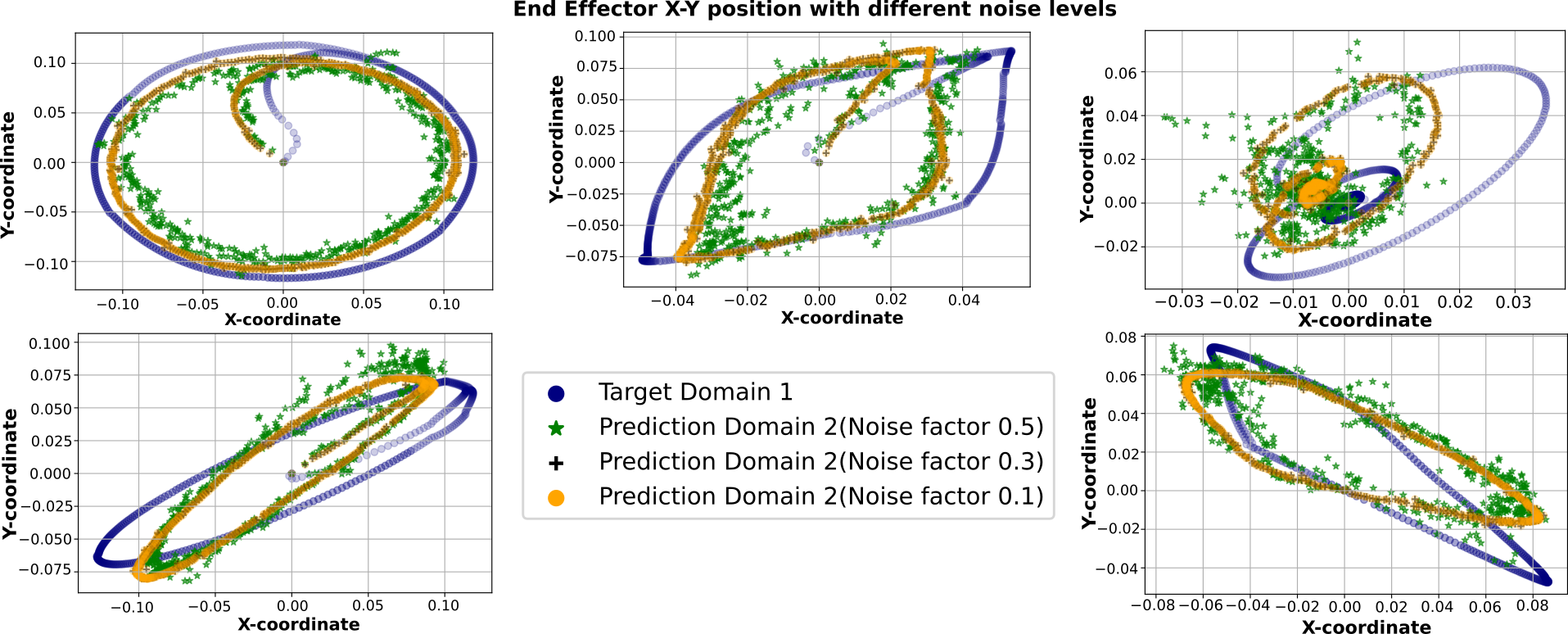}
	\caption{Impact of noise on the predictions of the generator $G_c$. We add gaussian noise of mean 0 and standard deviation 0.1, 0.3 and 0.5 to the original pressure of the source domain $\tau_{D_s}$}
	\label{fig:noise_all}
\end{figure*}
\begin{figure*}
	\centering
	\includegraphics[]{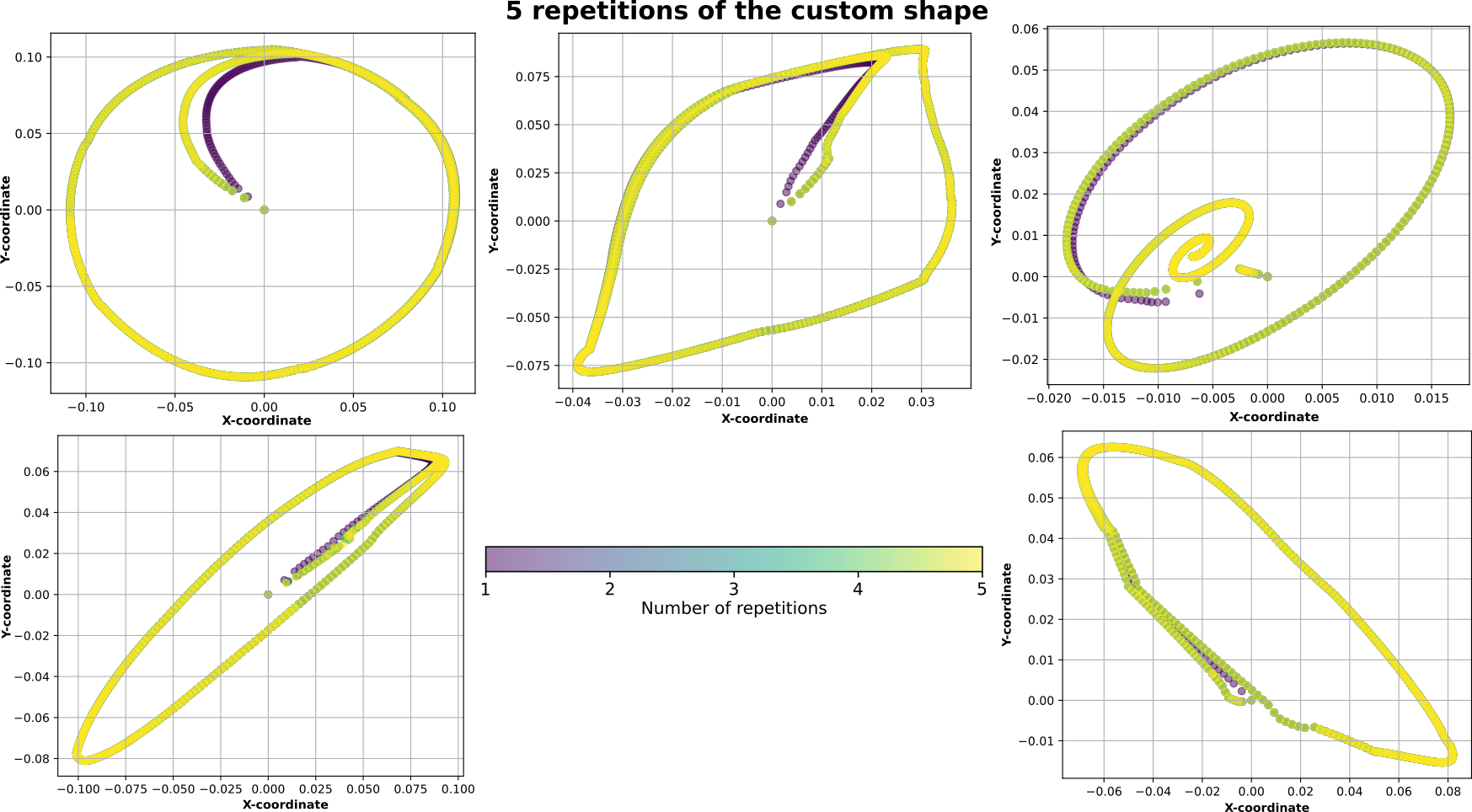}
	\caption{Periodicity experiment to check the stochasticity present in the generator $G_c$ predictions. All the shapes are generated by sending the actuation signals for 60 seconds with the soft robot operating at 10 Hz.}
	\label{fig:repeat_all}
\end{figure*}

\begin{table*}[]
\caption{End-effector position error $X_{err}$ (in cm) and orientation error $\phi_{err}$ (in radians) during trajectory-tracking experiments for the proposed algorithm as well as the benchmark algorithm.}
\label{pred_table}
\begin{center}
\begin{tabular}{*5c}
\hline
 & \multicolumn{2}{c}{\textbf{CCGAN-GP (Our)}}  & \multicolumn{2}{c}{\textbf{WCGAN-GP}} \\[1mm]
\hline
\textbf{Shape} & $X_{err}$  & $\phi_{err}$ & $X_{err}$  & $\phi_{err}$ \\[1mm] 
\hline
Circle & 2.98$_{\pm 0.29}$ & 0.2758$_{\pm 0.061}$ & 7.14$_{\pm 1.94}$ & 0.636$_{\pm 0.098}$ \\[1mm] 
Rectangle & 1.56$_{\pm 0.60}$ & 0.126$_{\pm 0.078}$ & 6.82$_{\pm 0.85}$ & 0.631$_{\pm 0.036}$ \\[1mm]
Damp circle & 1.19$_{\pm 0.48}$ & 0.101$_{\pm 0.043}$ & 8.78$_{\pm 2.58}$ & 0.83$_{\pm 0.210}$ \\[1mm] 
Ellipse & 1.56$_{\pm 0.35}$ & 0.126$_{\pm 0.069}$ & 7.22$_{\pm 0.43}$ & 0.717$_{\pm 0.081}$ \\[1mm] 
Kite & 2.37$_{\pm 0.87}$ & 0.19$_{\pm 0.083}$ & 9.02$_{\pm 3.17}$ & 0.81$_{\pm 0.255}$ \\[1mm] 
\hline
\end{tabular}
\end{center}
\end{table*}

\subsection{Training \& Evaluation Process}\label{train_gan}
The complete training procedure includes the adversarial training of two generator networks $(G_s, G_c)$ and two critic networks $(C_s, C_c)$ in a closed loop with the forward model $(F_c)$ to favour quick convergence and fast learning as shown in Figure \ref{fig:architecture}. The input to $G_c$ is the actuation of domain 1 ($\tau^i_{D_s}$) conditioned by the corresponding end effector pose ($X^{D_s}_i, \phi^{D_s}_i$) whereas the input to $G_s$ is the actuation of domain 2 ($\tau^i_{D_c}$) and the end effector pose ($X^{D_c}_i, \phi^{D_c}_i$). The output of $G_s$ further goes to the critic $C_s$ and the output of $G_c$ goes to $C_c$ to train the critic networks.

In addition, the output of $G_c$ also goes as an input to $G_s$ and vice-versa to create a cyclic training. Since we are mainly focused on the transfer of robotic skills from domain 1 to domain 2, so we just integrate the forward model of domain 2 $(F_c)$ in the training loop. We extract the predicted actuation of $G_c$ for domain 2 and pass it through the $F_c$ model to predict the end effector pose for that respective actuation value. The predicted pose in domain 2 and the original pose in domain 1 are then compared to calculate the feedback error. We want our $G_c$ model to make predictions such that it can replicate the same trajectories of $D_s$ in $D_c$ which is a more constrained domain. 

To address the constraints of $D_c$, the $G_c$ model is equipped with an adaptive time stretching phase as mentioned before. During this phase, the generator model estimates the amount of interpolation ($\lambda_d$) required for the predicted actuation to replicate the trajectory in the target domain with minimal error in the end effector pose. To determine the optimal value of $(\lambda)_d$, the semi-trained model enters an intermediate test phase where it makes predictions on validation data and calculates the error between the original target domain's end effector pose and the pose generated using the $G_C$'s prediction. The $G_c$ model adjusts $(\lambda)_d$ after every $N$ steps (a hyperparameter) and selects the value that minimizes the error.

The architecture is trained for 550 epochs with a batch size of $10$ and a learning rate of $10^{-4}$ for the generator and $2$ x $10^{-4}$ for the discriminator networks. Once the training is completed the network proceeds to the test phase, during the test phase we integrate the original deterministic simulator with our GAN framework instead of using the LSTM-based approximate forward model. We don't integrate it during the training phase due to its slow sequential structure and lack of parallelization capabilities as mentioned before. The complete training process takes around 12 hours 32 minutes on a computer equipped with AMD Ryzen 9 CPU (NVIDIA RTX 3060 GPU) running Ubuntu 22.04. The value of the hyperparameters are calculated using Optuna framework \cite{akiba2019optuna} which uses Bayesian parameter search over a provided range of values.

In the test phase, for any arbitrary trajectory, actuation data from the standard source domain ($D_s$) and the associated end-effector pose are passed through $G_c$, which then generates the actuation pressures required to reproduce the same trajectory in the target domain ($D_c$) with minimal pose error.

\section{Evaluation Criteria}

To evaluate the performance of our proposed framework, we simulate five custom shapes in $D_s$ by providing appropriate values to the three pressure chambers to perform a trajectory tracking experiment. Specifically, the end effector of the soft arm is directed to trace a circle, rectangle, damped circle, kite, and ellipse. These shapes are generated by sending 60-second pressure commands to the three chambers of the soft robot for domain 1. The pressure values for all three chambers are generated by incorporating an additional phase term. For example, in the case of the circle (as shown in Figure \ref{fig:circle_rect}), a sinusoidal signal with phases 0, 120, and 240 degrees are applied to the first, second, and third valves of the soft arm. The period of the circle is 30 seconds, ensuring it completes two full revolutions within the 60-second time frame. By altering the type of actuation signal (e.g., sinusoidal, ramp, staircase, etc.) as well as the phase angle, we can generate different and more complex shapes in the task space. The trajectories are displayed only in the \(XY\) plane, as visualizing the shapes in a 3D plot makes the analysis more challenging.

The generator model for domain 2, \( G_c \), is responsible for generating the pressure values for the three chambers, enabling the robot to replicate the same shapes in \( D_c \) as in \( D_s \). To evaluate the quality of the generated shapes in \( D_c \), we define two metrics for assessing the position and orientation: (1) the L2 norm between the predicted position values and the corresponding task-space positions from the source domain, and (2) the cosine similarity between the predicted and source domain orientation values of the robot’s end effector, as formulated in the equations below. Additionally, to assess the stability of our algorithm, we conduct an adversarial noise test, where small arbitrary noise is added to the base actuation signals from \( D_s \) before passing them to \( G_c \). Furthermore, we perform a periodicity experiment to analyze the stochasticity in the predicted actuation values within \( D_c \).

\begin{equation}
    \label{cosine_error}
    \phi_{error} = \arccos\big({\frac{\phi^P_{D_c}}{\phi^T_{D_s}}\big)}
\end{equation}
\begin{equation}
    \label{l2_error}
    X_{error} = \left \| X^P_{D_c} - X^T_{D_s} \right\|_2
\end{equation}

where $X^P_{D_c}, \phi^P_{D_c} = F_c(\tau^P_{D_c})$ and $\tau^P_{D_c} = G_c(\tau_{D_s}, X_{D_s}, \phi_{D_s})$ $ \forall $ $\tau_{D_s}, X_{D_s}, \phi_{D_s} \in D_s$.

Futhermore, we train a benchmark model to compare with our proposed algorithm. Specifically, we implement a wasserstein conditional GAN with a gradient penalty term (WCGAN-GP), a feedback loss, and the adaptive time-stretching mechanism, but without the cycle loss. This architecture consists of a single generator and a single critic network, trained adversarially in a manner similar to CCGAN. However, unlike our proposed approach, the WCGAN model is trained solely on the domain 2 data, as it includes only a single generator and critic network. The training process follows the same parameters and protocols as those used for the proposed method.

\section{Results \& Discussion}

\subsection{Domain Translation}
To evaluate the domain translation capability of the proposed CCGAN-WP algorithm as well as the benchmark algorithm, we perform a dynamic trajectory tracking experiment. Specifically, the actuation sequences corresponding to various shapes from the source domain, along with their associated end-effector poses, are input to the generator model $G_C$. This model translates these actuations into the target domain $D_c$. The resulting actuations are then passed through the forward model of $D_c$, and the tracking error is computed by comparing the predicted trajectory with the original ground truth trajectory.

Figure \ref{fig:circle_rect}, and \ref{fig:circle_rect2} (right), illustrates the end effector's position with respect to time for all three axes, whereas the plots on the left depicts the trajectory traced by the end effector of the robot in the X-Y plane. The plot in green demonstrates that the generator $G_c$ is able to transfer the actuation of the source domain to the target domain with acceptable error in the task space. The plot in blue denotes the original XYZ values of the end effector in the source domain. The plot in maroon, highlights the shape traced when the actuations from $D_S$ are directly passed through the forward model $F_C$, without any domain translation. In red are the predictions made by the benchmark model. It can be seen that the proposed model performs better than the benchmark model in capturing both the position as well as the orientation of the effector for the different shapes. Performance is notably better for circular and ellipsoidal shapes, while for shapes with sharp corners, such as rectangles or kites, the model struggles to retain the desired form. This is likely due to the limited capability of the model to capture non-sinusoidal signals. Table \ref{pred_table} summarizes the overall position error (in cm) and orientation error (in radians) for all five shapes. It can be seen that the error of our proposed approach is below the acceptable error limit (3 cm, 0.3 radians). The experiment is conducted three times with different random seeds, and the error is computed using equations  \ref{cosine_error} and \ref{l2_error} for each data point, followed by calculating the mean and standard deviation for the respective shapes.

\subsection{Noise Test}
Figure \ref{fig:noise_all} illustrates the predictions of the generator \( G_c \) when subjected to noisy input signals. As previously mentioned, we introduce slight corruption to the input actuation signals (\( \tau^{D_s} \)) by adding arbitrary gaussian noise with a mean of 0 and standard deviation of 0.1 (depicted with orange circles), 0.3 (black plus signs), and 0.5 (green stars). The results indicate that the model generally maintains robust predictions despite the presence of external noise. However, for the damped circular trajectory, the predictions exhibit significant deviations at the highest noise level. A possible explanation for this behavior is the inherent damping effect of the trajectory, which becomes more scattered under extreme noise conditions.

\subsection{Periodicity Test}

To assess the level of stochasticity in the generator model's predictions, we repeat the evaluation phase five times for each of the five different shapes. Before each repetition, the model is reset to its initial rest position and then traverses the desired shape. Figure \ref{fig:repeat_all} illustrates the predictions generated by \( G_c \) across multiple repetitions. Notably, the model consistently produces stable predictions across all repetitions, demonstrating its reliability.


\section{Conclusion}
In this paper, we introduce a method for transferring system dynamics learned in a source domain (\(D_s\)) to a target domain (\(D_c\)), where the environment exhibits a tenfold increase in viscosity for a soft robotic arm. This technique applies in soft robotic scenarios where system dynamics shift or degrade over time. Our approach leverages a Cycle-GAN-based architecture, trained adversarially on data from both domains, incorporating four distinct loss functions. The model is trained using a simulated dataset consisting of actuation pressure and corresponding end-effector poses collected from 50 randomly generated trajectories per domain. 

The effectiveness of the proposed architecture is evaluated through trajectory-tracking experiments on five distinct shapes, comparing its performance against a baseline GAN model trained solely on babbling data from \(D_c\) without cyclic training. To further assess robustness, we introduce an adversarial noise test by adding random Gaussian noise to the actuation inputs of \(D_s\) before feeding them into \(G_c\). Additionally, a periodicity test examined the stochastic nature of the predictions generated by \(G_c\). To our knowledge, this is the first time simulations of soft robot control supported a domain translation study.

The present study also has limitations. First, our approach requires an extended training time primarily due to the sequential batch-wise data processing. Future work will mitigate this by parallelizing computations across trajectories to accelerate convergence. Another limitation is that our method has been validated only in simulation; as a next step, we plan to evaluate its performance on a real robotic system.

\bibliographystyle{ieeetr}
\bibliography{ccgan.bib}

\end{document}